\documentclass[journal]{IEEEtran}
\ifCLASSINFOpdf
   \usepackage[pdftex]{graphicx}
   \usepackage{caption}
\else
\fi
\usepackage{multirow}
\usepackage{textcomp}
\usepackage[flushleft]{threeparttable}

\begin{document}
%
\title{Robust Object Classification Approach using Spherical Harmonics}
%
%
%
\author{\uppercase{Ayman Mukhaimar},
\uppercase{Ruwan Tennakoon, Chow Yin Lai, Reza Hoseinnezhad, and Alireza Bab-Hadiashar},
}
\author{\IEEEauthorblockN{Ayman Mukhaimar\IEEEauthorrefmark{1},
Ruwan Tennakoon\IEEEauthorrefmark{2},
Chow Yin Lai\IEEEauthorrefmark{3}, 
Reza Hoseinnezhad\IEEEauthorrefmark{1} and
Alireza Bab-Hadiashar\IEEEauthorrefmark{1}} \\
    \IEEEauthorblockA{\IEEEauthorrefmark{1}School of Engineering, 
RMIT University, Melbourne, Australia}\\
\IEEEauthorblockA{\IEEEauthorrefmark{2}School of Science,
RMIT University, Melbourne, Australia}\\
\IEEEauthorblockA{\IEEEauthorrefmark{3} Department of Electronic \& Electrical Engineering, University College London, UK
}
\thanks{This work was supported by ....}
\thanks{Manuscript received mmmm dd, yyyy; revised mmmm dd, yyyy.}}

\maketitle

\begin{abstract}
In this paper, we present a robust spherical harmonics approach for the classification of point cloud-based objects. Spherical harmonics have been used for classification over the years, with several frameworks existing in the literature. These approaches use variety of spherical harmonics based descriptors to classify objects. We first investigated these frameworks robustness against data augmentation, such as outliers and noise, as it has not been studied before. Then we propose a spherical convolution neural network framework for robust object classification. The proposed framework uses the voxel grid of concentric spheres to learn features over the unit ball. Our proposed model learn features that are less sensitive to data augmentation due to the selected sampling strategy and the designed convolution operation. We tested our proposed model against several types of data augmentation, such as noise and outliers. Our results show that the proposed model outperforms the state of art networks in terms of robustness to data augmentation.

\end{abstract}


%
\IEEEpeerreviewmaketitle

\section{Introduction}
%
%
%
%
\IEEEPARstart{T}{he} recent success and popularity of Convolutions Neural Networks (CNN) for many computer vision applications have inspired researchers to use those for 3D model classification as well \cite{qi2017pointnet,zhou2018voxelnet,su2015multi}. To exploit the potential of deep networks for this application, different representations of 3D data such as kd-tree \cite{klokov2017escape}, dynamic graphs \cite{wang2019dynamic}, and most recently, spherical harmonics \cite{cohen2018spherical,esteves2018learning,perraudin2019deepsphere,poulenard2019effective} have been proposed. Spherical harmonics is a representation that have attracted significant interest in a wide range of applications including matching and retrieval  \cite{kazhdan2003rotation,wang20163d}, lighting \cite{green2003spherical}, and surface completion \cite{nortje2015spherical}. They attain several favourable characteristics for working with 3D space, such as their basis are defined on the surface of the sphere (volumetric) and are rotation equivariant. In addition, they have shown to provide compact shape descriptors compared to other types of descriptors \cite{kazhdan2003rotation,bulow2001surface}. 

The use of CNNs with spherical harmonics has had major success in several recent papers for shape classification \cite{cohen2018spherical,esteves2018learning,poulenard2019effective}, retrieval \cite{esteves2018learning} and alignment \cite{esteves2018learning}. Unlike conventional approaches that use CNNs in regular Euclidean domains, spherical harmonics CNNs (SCNNs) apply convolutions in SO(3) Fourier space, learning features that are SO(3)–equivariant. Interestingly, spherical CNNs have shown to have fewer parameters \cite{cohen2018spherical} and faster training due to the reduction in the dimensionality of the spherical harmonics shape descriptors. 

Frameworks that use spherical harmonics CNNs can be divided into two groups: Point-based SCNN that extract features based on point maps or pairwise relations \cite{poulenard2019effective,perraudin2019deepsphere} and the other group that uses spherical harmonics convolution on images casted on the sphere \cite{cohen2018spherical,esteves2018learning}. Approaches that use images casted on the sphere still need to include rotation in the training stage as casted images change with rotation. Nevertheless, it has been shown that spherical convolutional neural networks outperform conventional CNNs, even when both are trained with rotations \cite{cohen2018spherical,esteves2018learning}.

 While spherical CNNs achieved impressive performance in several 3D tasks, their robustness to uncertainty in three-dimensional data has not been explored. Point clouds of 3D models produced by either 3D scanners or multi-view images are often imperfect and contain outliers. As such, development of robust classification frameworks that can deal with such inaccuracies is of interest. Our experiments with using the state-of-art SCNNs with data containing outliers showed that SCNN approaches that are based on point-map features often fail to classify shapes due to outliers distorting the map graph. Also, methods that use image casting typically find the farthest point of intersection between the casted rays and the 3D shape; hence there is a high chance that outlier points are selected as the farthest points. Surprisingly, even when median was used instead of the farthest point, these approaches still often fail to classify shapes.

\begin{figure*}[t]
    \centering
    \includegraphics[scale=.53]{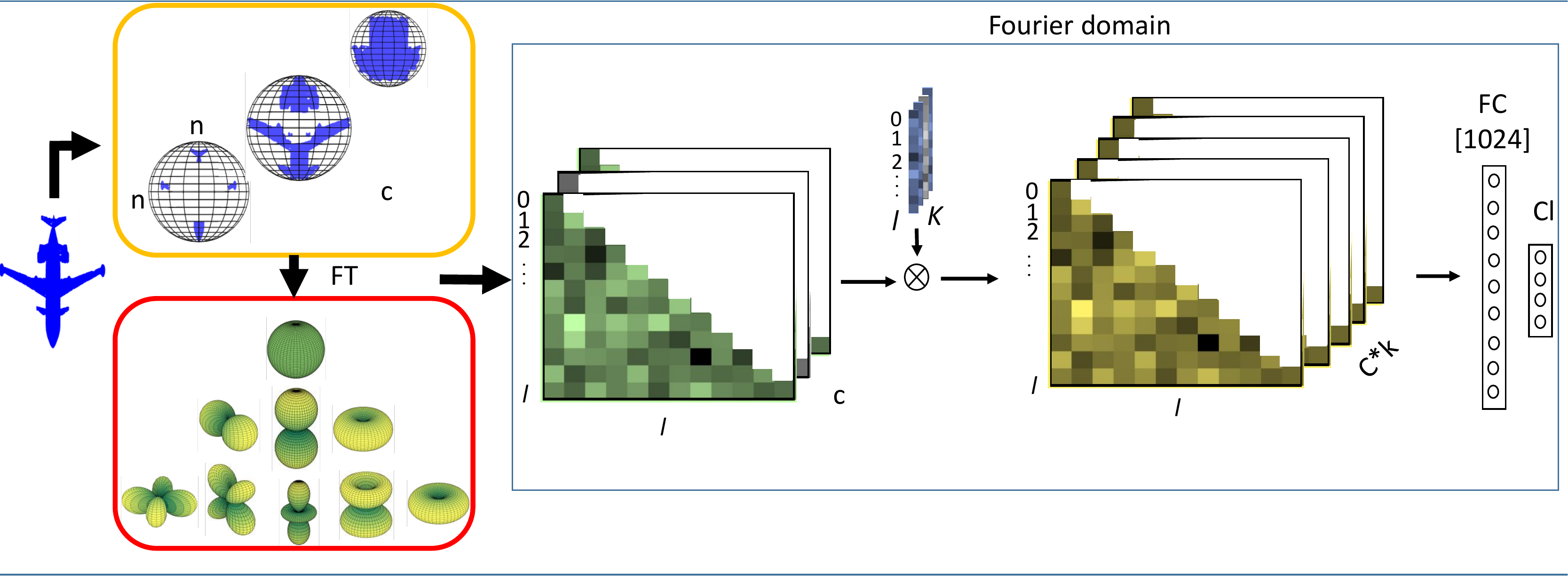}
    \caption{The Proposed Spherical CNN framework. First we sample the shape with $c$ concentric spheres and $n \times n$ grid resolution (seen in orange box). Next we apply Fourier transform (FT) (seen in red box) on the spherical signal and get the basis coefficients up to order \textit{l}. Then we apply the spherical convolution operation where k, is the number of filters. Finally, we feed the spectrum into a fully connected layers (FC) and a classification layer (Cl).}
    \label{fig:fig1}
\end{figure*}
 
 In this paper, we investigate the robustness of spherical harmonics to data inaccuracies. To start, we explain the spherical harmonics descriptors and the common sampling strategies used in the literature. We then show that using concentric spheres with density occupancy grids provide the highest robustness against data augmentation. The use of concentric spheres generates uniform voxel grids, which have shown to be robust to data augmentation \cite{mukhaimar2019comparative,Riegler_2017_CVPR}. We also propose to use the magnitude of each specific spherical component for shape classification and show that it produces better robustness compared to using the combined magnitudes of different components in each order. In particular, we show that a simple classifier (i.e. fully connected neural network) with the previously mentioned spherical harmonics descriptors and sampling strategy is robust to high levels of data augmentation. Using the above knowledge and the inspiration from the recent success of spherical CNNs approaches ~\cite{cohen2018spherical,esteves2018learning}, we propose a robust spherical convolutional neural network framework (called RSCNN) that is able to deal with different types of uncertainty inherent in three-dimensional data measurement. The framework is outlined in Figure 1. In our approach, the entire spherical convolution operation is performed in the Fourier domain (we show that using the inverse transform reduces robustness to data augmentation), and the sampling is performed over concentric spheres. 

Our proposed approach uses voxel grids, similar to 3D CNNs~\cite{Riegler_2017_CVPR}, that have shown to be robust to the influence of outliers. However, we are using spherical grid that has been shown to exhibit better rotation equivarience compared to euclidean grids \cite{boomsma2017spherical} and using spherical CNNs that have less trainable parameters \cite{cohen2018spherical}. Unlike previous approaches \cite{qi2017pointnet,qi2017pointnet++,su2015multi}, the applied spherical convolution operation is simply multiplying the filter kernels by the spherical harmonic coefficients, hence, the convolution operation does not disrupt the input signal through the use of a pooling operation or grid altering. Also, we will show the output features produced by the convolution operation are highly robust. 


Our key contributions in this paper are as follows:
\begin{itemize}
\item We investigate the robustness of spherical CNNs and spherical harmonics as it has not discussed before.
\item We propose a new spherical CNN framework that is significantly more robust than previous spherical CNNs approaches.
\item We demonstrate the efficiency and accuracy of our method on shape classification with the presence of several types of data inaccuracies, and we show that our framework significantly outperform all previous approaches.
\end{itemize}

The rest of this paper is structured as follows. An in-depth literature review is provided in section \ref{sec:Related work}, while preliminaries required to explain spherical harmonics are provided in section \ref{sec:Preliminaries}. A detailed discussion of the proposed model is presented in section \ref{sec:Method}. Description of an extensive set of experiments is included in section \ref{sec:Experiments} as well as a comparison with other the state of the art 3D shape classification (including other spherical harmonics) methods. We also examine the robustness of different spherical harmonics structures and descriptors in section \ref{sec:ABLATION STUDY}. Section \ref{sec:Conclusion} concludes the paper.

\section{Related work}
\label{sec:Related work}
Spherical harmonics have been used for 3D shape classification for many years. Early classification frameworks used spherical harmonic coefficients as shape descriptors \cite{kazhdan2003rotation}. Later, with the use of spherical convolutions, classifier networks were empowered to learn descriptive features of objects. We study both approaches in terms of their performance under data augmentation. 

The spherical CNNs proposed in \cite{cohen2018spherical,esteves2018learning}, for spherical signals defined on the surface of a sphere, addresses the rotation equavarience using convolutions on the SO3 rotation group. In \cite{cohen2018spherical}, the input spherical signal is convolved with S2 convolution to produce feature maps on SO3, followed by an SO3 convolution, 
while in \cite{esteves2018learning}, spherical convolutions were used. Steerable filters  \cite{worrall2017harmonic,weiler2018learning,weiler20183d} were used to achieve rotation equivariance, where filters were restricted to the from of complex circular harmonics \cite{worrall2017harmonic}, or complex valued steerable kernels were used \cite{weiler2018learning}. In \cite{poulenard2019effective}, a network called Sphnet is designed to apply spherical convolution on volumetric functions \cite{atzmon2018point} generated using extension operators applied on point cloud data. Unlike previous approaches, spherical convolution is applied on point clouds instead of spherical voxel grid, resulting in better rotation equivariant. In another work, a network called Deepsphere \cite{perraudin2019deepsphere},  spherical CNNs are used on graph represented shapes, where shapes are projected onto the sphere using HEALPix sampling, in which the relations between the pixels of the sphere build the graph. The graph is then represented by the Laplacian equation, which is solved using spherical CNNs. Ramasinghe et al. \cite{Ramasinghe} investigated the use of radial component in spherical convolutions instead of using spherical convolutions on the sphere surface. They proposed a volumetric convolution operation that was derived from Zernike polynomials. Their results show that the use of volumetric convolution provides better performance by capturing depth features inside the unit ball. Spherical signals have also been used in conjunction with conventional CNNs by \cite{boomsma2017spherical,su2017learning,coors2018spherenet} to achieve better rotational equivariance than signals on euclidean space, but unlike spherical CNNs, their CNN operations are not equivariant. You et al. \cite{you2018prin} used concentricity to sample 3D models with a sampling strategy that has better robustness to rotation.
While previous approaches used spherical harmonics to build rotation equavarient neural networks, our focus is on building a robust spherical harmonics structure. As such, our choice of representation is not restricted to spherical harmonics descriptors that are rotation equavarient. 

In terms of recent deep learning approaches for 3D shape classification, 3D CNNs have been used with voxel based 3D models \cite{maturana2015voxnet,wu20153d,zhou2018voxelnet,Riegler_2017_CVPR} using several occupancy grids \cite{maturana2015voxnet}. Such a representation has shown to be robust to data augmentation \cite{Riegler_2017_CVPR} while some implementations (e.g.~\cite{brock2016generative}) have achieved very high classification accuracy for clean objects (using ModelNet40). 
Another approach is to use 2D CNN on images of the 3D mesh/CAD objects rendered from different orientations \cite{su2015multi,johns2016pairwise,wang2019dominant}. The rendered images are usually fed into separate 2D CNN layers, a pooling layer follow these layers to aggregate their information. These methods take advantage of existing pertained models to achieve high classification accuracy. When testing MVCNN \cite{su2015multi}, the classification accuracy was heavily affected by data augmentation, especially outliers. Another approach uses unsorted and unprocessed point clouds directly as an input to the network layers \cite{qi2017pointnet,qi2017pointnet++}. These approaches use a max-pooling layer that was tested to be robust to point dropout and noise \cite{mukhaimar2019comparative}. However, when tested with outliers, their performance was significantly affected. Another approach is to build upon relations between points \cite{klokov2017escape}; for such methods, the existence of outliers completely changes the distance graph and causes such an approach to fail.

In terms of robust classification frameworks that exist in the literature, Pl-net3D \cite{mukhaimar2019pl} decompose shapes into planar segments and classify objects based on the segments information. DDN~\cite{gould2019deep} proposes an end-to-end learnable layer that enables optimization techniques to be implemented in conventional deep learning frameworks. An m-estimators based robust pooling was proposed instead of max pooling used in conventional CNNs. Our approach shows better robustness to data argumentation while involves less computation. 

\section{Preliminaries}  
\label{sec:Preliminaries}

In this section, we review the theory of spherical harmonics along with their associated descriptors that are used for classification tasks. In addition, we review the theory of convolution operations applied to spherical harmonics.

\subsection{Spherical harmonics}

Spherical harmonics are a complete set of orthonormal basis functions that are defined on the surface of unit sphere $S2$ as: 

\begin{equation}
Y_{l}^{m}(\theta,\varphi)=\sqrt{{(2l+1)(l-m)!\over 4\pi(l+m)!}}P_{l}^{m}(\cos\theta)e^{im\varphi}
\label{equ:sh}
\end{equation}

where $P_{lm}(x)$  is the associated Legendre polynomial, $l$ is the degree of the frequency and $m$ is the order of every frequency $(l\geq 0, m\leq\vert l\vert). ~\theta\in[0,\pi],   ~\varphi\in[0,2\pi]$ denote the latitude and longitude, respectively. Any spherical function $f(\theta,\varphi)$ defined on unit sphere $S2$ can be estimated by the linear combination of these basis functions:
\begin{equation}
  f(\theta,\varphi) = \sum\limits_{l=0}^{\infty}\sum\limits_{m=-l}^{\l}\hat{f}_{lm}Y_{l}^{m}(\theta,\varphi)
  \label{equ:fun}
\end{equation}
  
where $\hat{f}_{lm}$ denotes the Fourier coefficient found from:
 \begin{equation} 
  \hat{f}(l,m) = \int_{S^{2}}f(\theta,\varphi)\bar{Y_{l}}^{m}(\theta, \varphi)d\varpi
  \label{equ:coef}
\end{equation}

The following descriptors implies that any spherical function can be described in terms of the amount of energy $|\hat{f}|$ it contains at every frequency: 
 \begin{equation} 
  F1 =(|\hat{f}_{0,0}|,|\hat{f}_{1,0}|,|\hat{f}_{1,1}|,...|\hat{f}_{lm}|)
  \label{equ:desc1}
\end{equation}


 \begin{equation} 
  F2 =(|\hat{f}_{0}|,|\hat{f}_{1}|,|\hat{f}_{2}|,...|\hat{f}_{l}|),~where~  \hat{f}_{l} =\sqrt{\sum\limits_{m=-l}^{\l}\hat{f}_{lm}^2}
  \label{equ:desc2}
\end{equation}

Both of these descriptor vectors have been used for classification of shapes \cite{kazhdan2003rotation,wang20163d}. However, only the second descriptor is rotation equavareint while the first one carries more shape information. We have investigated the use of both descriptors for shape classification and the result is provided in section~\ref{sec:Method}.

\subsection{Spherical convolution}

If we have a function $f$ with its Fourier coefficients $\hat{f}$ found from equation~(\ref{equ:coef}), and another function or kernel h with its Fourier coefficients $\hat{h}$, then the convolution operation in spherical harmonic domain is equal to the multiplication of both functions Fourier coefficients as shown below \cite{driscoll1994computing}:

 \begin{equation} 
 (f * h)_{l}^{m} =  \sqrt {{4\pi} \over {2\ell +1}} \hat{f}_{\ell}^{m} \hat{h} _{\ell}^{0}.  
   \label{equ:fh}
\end{equation}

Here, the convolution at degree $l$ and order $m$ is obtained by multiplying of the coefficient $\hat{f}_{\ell}^{m}$ with the zonal filter kernel $\hat{h}_{\ell}^{0}$. The inverse transform is also achieved by summing over all l values: 

\begin{equation}
  (f * h)(\theta,\varphi) = \sum\limits_{l=0}^{\infty}\sum\limits_{\vert m\vert \leq l}(f * h)_{l}^{m}Y_{l}^{m}(\theta,\varphi).
  \label{equ:fh1}
\end{equation}

\section{Method}
\label{sec:Method}
To facilitate the description of our solution space, we first introduce the shape classification problem and the effect of outliers, followed by depiction of our proposed solution architecture (as shown in Figure~\ref{fig:fig1}). The proposed solution starts from sampling the input spherical signal, then moves to convolution operations, and finally performs classification.  

\subsection{Problem statement}

The point clouds of 3D models produced by either 3D scanners or multi view images are often imperfect and contain outliers. To be able to mitigate the effect of outliers, we need to develop ways to model the data. Our primary objective for using spherical harmonics is to achieve robust classification rather than rotation equavarience. As such, we constrain the object rotations to a rotation around z-axis similar to most contemporary approaches including \cite{qi2017pointnet,zhou2018voxelnet,wang2019dynamic}.  

\subsection{Sampling on concentric spheres}

The first step in using the spherical harmonics for modelling an object is to sample the input signal, which is referred to as $f(\phi,\theta)$ in equation~(\ref{equ:coef}). Two types of sampling are used in literature \cite{kazhdan2003rotation}: Sampling over the concentric spheres, and sampling over the sphere surface. Our results show that the method \cite{esteves2018learning} that use image casting fails to mitigate the effect of outliers. Therefore, the second sampling strategy is adopted in this paper. The generated grids for both types of sampling change with rotation and therefore, rotation needs to be included during training.

We generate a spherical voxel grid that consists of $c$ concentric spheres with $n \times n$ grid resolution for each concentric sphere. The generated spherical voxel grid allows sampling over the unit ball $(S3)$, with each voxel being represented by $(r,\theta,\phi)$ where ($r \,\to\, 0:c.$ $ \theta,\phi \,\to\, 0:n)$. We distribute the given 3D shape over the grid and we keep record of number of points inside each voxel and produce a density occupancy grid. The use of such occupancy grid is expected to provide reliable estimates in the presence of outliers. We compare occupancy grids in the next section.

\subsection{Classification with Fourier coefficients}

The robustness of the two descriptors represented by equation~(\ref{equ:desc1}) and equation~(\ref{equ:desc2}) for object classification are compared in this section by feeding each of them to a fully connected neural network. We tested their robustness against: Gaussian noise with 0.10 standard deviations, uniformly scattered outliers with 50\% percentage, and 80\% Random point dropout. More about those tests are detailed in the experiment section. The results are shown in table~\ref{Tab:t1}. The results in table~\ref{Tab:t1} also show the robustness of the sampling strategies and the occupancy grids mentioned in the previous section. The results show that the used sampling and the density occupancy grid provides a high degree of robustness to outliers. In addition, these results also show that the descriptor in equation~(\ref{equ:desc1}) F1 provides higher classification accuracy than using the descriptor in equation~(\ref{equ:desc2}) F2 as the first one carries more shape information.

\begin{table}[]
 \centering
\begin{threeparttable}
\caption{Classification accuracy results for objects augmented with noise, missing points, and outliers.}
\label{Tab:t1}
 \fontsize{10}{12}\selectfont 
\begin{tabular}{|l|c|c|c|c|}
\hline
\multicolumn{1}{|c|}{\begin{tabular}[c]{@{}c@{}}occupancy   \\ grid\end{tabular}} & clean  & \begin{tabular}[c]{@{}c@{}} random \\ dropout\end{tabular} & \begin{tabular}[c]{@{}c@{}} noise\end{tabular} & \begin{tabular}[c]{@{}c@{}} outliers\end{tabular} \\ \hline \hline
binary                                                                            & \textbf{0.79} & 0.34                                                           & 0.24                                              & 0.14                                                 \\ \hline
density+F1                                                                        & 0.78   & \textbf{0.75}                                                           & \textbf{0.37}                                            & \textbf{0.50}                                                     \\ \hline
density+F2                                                                        & 0.68  & 0.68                                                         & 0.3                                                 & 0.24                                                 \\ \hline
\end{tabular}
    \begin{tablenotes}
      \small
      \item The dataset details are shown in section~\ref{sec:Experiments}.
    \end{tablenotes}
  \end{threeparttable}
  
\end{table}



\subsection{Implementation of Spherical convolution}

We propose to apply the spherical convolution on 3D models that are decomposed into concentric spheres. The use of concentric spheres generates a uniform spherical voxel grid that enables the spherical convolution neural network to learn features over the unit ball (as appose to only learning over the unit sphere). We use separate convolution operation at each concentric sphere to allow our network to learn features relevant to that sphere. To achieve the spherical convolution, we use equation~(\ref{equ:fh}) in which the learned kernel is a zonal ($m=0$) filter h with dimension of $l*c$ , where $l$ is the frequency degree, and $c$ is the number of concentric spheres. Similar to \cite{esteves2018learning}, we parameterize the kernel filters in the spectral domain. No inverse Fourier transform is applied after the spherical convolution. Therefore, our convolution operation is entirely in the spectral domain, which reduces the convolution computation time. Our results show that applying inverse Fourier transform (IFT) diminishes the robustness to outliers as the overall accuracy reduces by almost 15 percent. This can be related to equation~(\ref{equ:fh1}), where for each $\theta$ and $\phi$, the output signal is calculated by summing the entire coefficients. Thus, if the coefficients are already altered by outliers, the output signal error will be magnified/accumulated due to this summation. A detail discussion on this topic is provided in the Appendix. 

Our experiments show that applying the convolution operation works well with augmented data and the network has been able to learn better features and be more discerning in terms of object classification. This is shown by applying t-sne~\cite{maaten2008visualizing} to clean and augmented data and the results are provided in the Appendix. As the application of convolutions on 3D voxel grids has shown to be robust to the influence of outliers \cite{mukhaimar2019comparative}, we would expect our method to exhibit a high degree of robustness to outliers as well.

Compared to previous approaches, unlike other networks such as pointnet~\cite{qi2017pointnet} where their max pooling chooses outliers as max, our proposed method does not use pooling or grid altering operations. Our proposed convolution operation can be described as follows: Let $x \in X$ be our input spherical coefficients at a given degree, the spherical convolution operation in equation~(\ref{equ:fh}) is simply $f(x_{i})=k \times x_{i}$, where $k$ is a constant calculated from the square root term of the same equation and the kernel value at that degree followed by the non-linearity operation. This mathematical operation does not alter the input signal and only assist with extracting better features in both clean and augmented data as seen from the t-sne results (provided in Appendix B).

Compared to 3D CNNs such as the octnet~\cite{Riegler_2017_CVPR}, spherical CNNs are rotation equavarient, which could help in increasing our performance. In addition, the use of spherical convolution have shown to have less trainable parameters, where one layer is enough to achieve good performance \cite{Ramasinghe}. 


\subsection{classification layer}

The returned feature map by the convolution operations represents the feature vector defined in equation 4. The map is then fed to a fully connected and a classification layers.  The feature vector in equation 5 could be used as well, however, it is less robust to data augmentation as shown in table 1. Although the feature vector defined in equation 4 is not rotation invariant, given that we are training with rotations, we would expect our network to learn rotations. 

\section{Experiments} \label{sec:Experiments}

The following experiments are carried out with $z$ rotation, only. We compare our framework with the state of the art published spherical convolution architectures, point cloud classification methods, and robust methods. We considered outliers, noise, and missing points as our types of data augmentation in this paper since corruption of point clouds with such inaccuracies are commonplace.   

\subsection{Dataset}

To test different methods, we use the benchmark dataset ModelNet40 \cite{qi2017pointnet++}. ModelNet40 dataset contains point clouds of 40 different shapes, with 9,843 point clouds for training, and 2468 for testing. Each object consists of 2048 points, which are normalized within the unit cube. We generated three artificial datasets from the ModelNet40 dataset for each of the three types of data augmentation. These datasets are only used during testing, while we only add noise and random rotations as our data augmentation during training.

\subsection{Architecture}

For the following experiments, we used one convolution layer having 16 output channels. We use prelu after the convolution operation as our non-linearity. The number of concentric spheres was chosen to be 7 with grid resolution of 64 by 64 of each sphere. The basis degree is selected to be 9. We compare different architectures and study the effect of changing the number of spheres in our ablation study.

\subsection{Training}

Similar to \cite{cohen2018spherical,esteves2018learning}, we also include rotation in the training data. The patch size is set to 16, learning rate varies from 0.001 to 0.00004, and number of epochs is set to 48. We used a TITAN Xp GPU, where only 350Mb of memory was used during training.\\

\subsection{Robustness to outliers}

Similar to \cite{mukhaimar2019pl}, we present two outlier scenarios generated with different mechanisms. In the first scenario, we test our model in the presence of scattered outliers: points uniformly distributed in the unit cube. In the second scenario, added outliers are grouped into clusters of ten or twenty points, which are uniformly distributed in the unit cube (similar to \cite{ning2018efficient}). The overall number of scattered points for this scenario are fixed to ten or twenty percent as shown in table~\ref{Tab:t4}. Points in each cluster are normally distributed with zero mean and standard deviations of 4\% and 6\%.

\begin{figure}[]
\centering
\includegraphics[scale=.33]{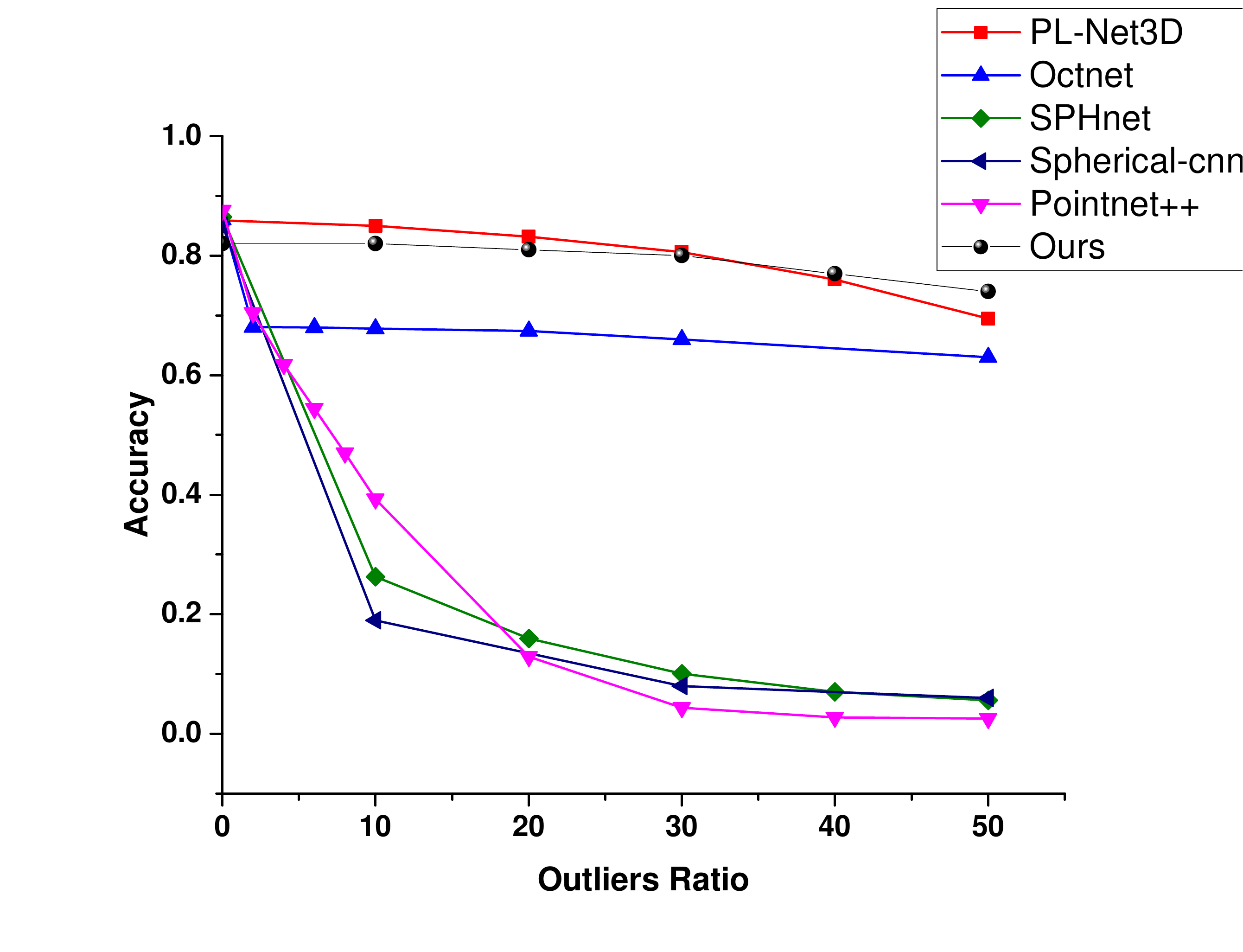}
\caption{Classification accuracy versus outliers.}
\label{fig_outl}
\end{figure}

Figure~\ref{fig_outl} and table~\ref{Tab:t4} show that our model is highly robust to the influences of outlier in both scenarios. The classification accuracy only drops by 6\% percent when half the data are outliers, while PL-net3D and other models drop by significantly higher margins. For Spherical-cnn~\cite{esteves2018learning}, even when we used the median aggregation for generating the unit sphere grid (instead of max aggregation), the network remains sensitive to the influences of the outliers. Similarly, SpH-net~\cite{poulenard2019effective} performs poorly when there were outliers as these outliers distort the distance graph. Results for using Spherical-cnn with our occupancy grid are examined in our ablation study.

\begin{table}[h]
\caption{Object classification results on clustered outliers}
\label{Tab:t4}
\centering
 \fontsize{10}{12}\selectfont 
\begin{tabular}{lllllllll}
\cline{1-4}
\multicolumn{1}{|c|}{\multirow{3}{*}{Method}} & \multicolumn{1}{c|}{10\%}    & \multicolumn{1}{c|}{10\%}    & \multicolumn{1}{c|}{20\%}    &  &  &  &  &  \\ 
\multicolumn{1}{|c|}{}                        & \multicolumn{1}{c|}{10p}     & \multicolumn{1}{c|}{10p}     & \multicolumn{1}{c|}{20p}     &  &  &  &  &  \\ 
\multicolumn{1}{|c|}{}                        & \multicolumn{1}{c|}{$\mathcal{N}$(0.04)} & \multicolumn{1}{c|}{$\mathcal{N}$(0.06)} & \multicolumn{1}{c|}{$\mathcal{N}$(0.04)} &  &  &  &  &  \\ \cline{1-4}
\multicolumn{1}{|c|}{Octnet}                  & \multicolumn{1}{c|}{0.47}    & \multicolumn{1}{c|}{0.48}    & \multicolumn{1}{c|}{0.37}    &  &  &  &  &  \\ \cline{1-4}
\multicolumn{1}{|c|}{PL-Net3D}                    & \multicolumn{1}{c|}{0.79}    & \multicolumn{1}{c|}{0.8}    & \multicolumn{1}{c|}{0.67}    &  &  &  &  &  \\

\cline{1-4}
\multicolumn{1}{|c|}{Ours}                    & \multicolumn{1}{c|}{\textbf{0.81}}    & \multicolumn{1}{c|}{\textbf{0.81}}    & \multicolumn{1}{c|}{\textbf{0.75}}    &  &  &  &  &  \\

\cline{1-4}
\multicolumn{4}{p{170pt}}{ 10\%: outliers percentage, 10p: 10 Points in each cluster.} 
\end{tabular}

\end{table}

\subsection{Robustness to noise}

\begin{figure}[]
\centering
\includegraphics[scale=.33]{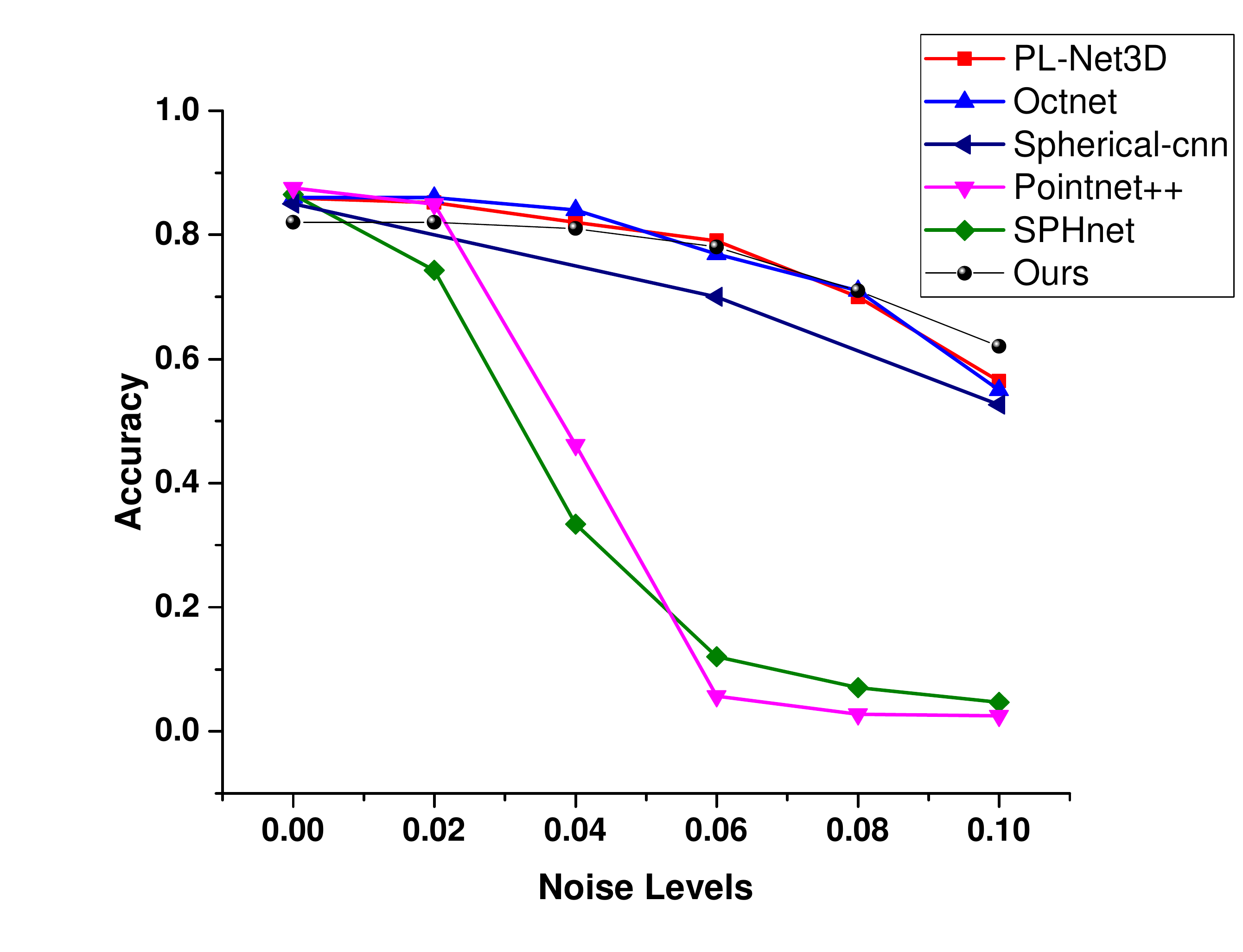}
\caption{Classification accuracy versus noise.}
\label{fig_noise}
\end{figure}

We simulated the effect of noise in point cloud data by perturbing points with zero mean, normally distributed values with standard deviations ranging from .02 to .10 as shown in Figure~\ref{fig_noise}. Compared to other models, our proposed model performance deteriorated the least (by around 18\%) for relatively large amount of noise (at 0.10 noise level). SPHnet was significantly affected by noise, while spherical-CNN performance was relatively much better than SPHnet.

\subsection{Robustness to missing points}

\begin{figure}[]
\centering
\includegraphics[scale=.33]{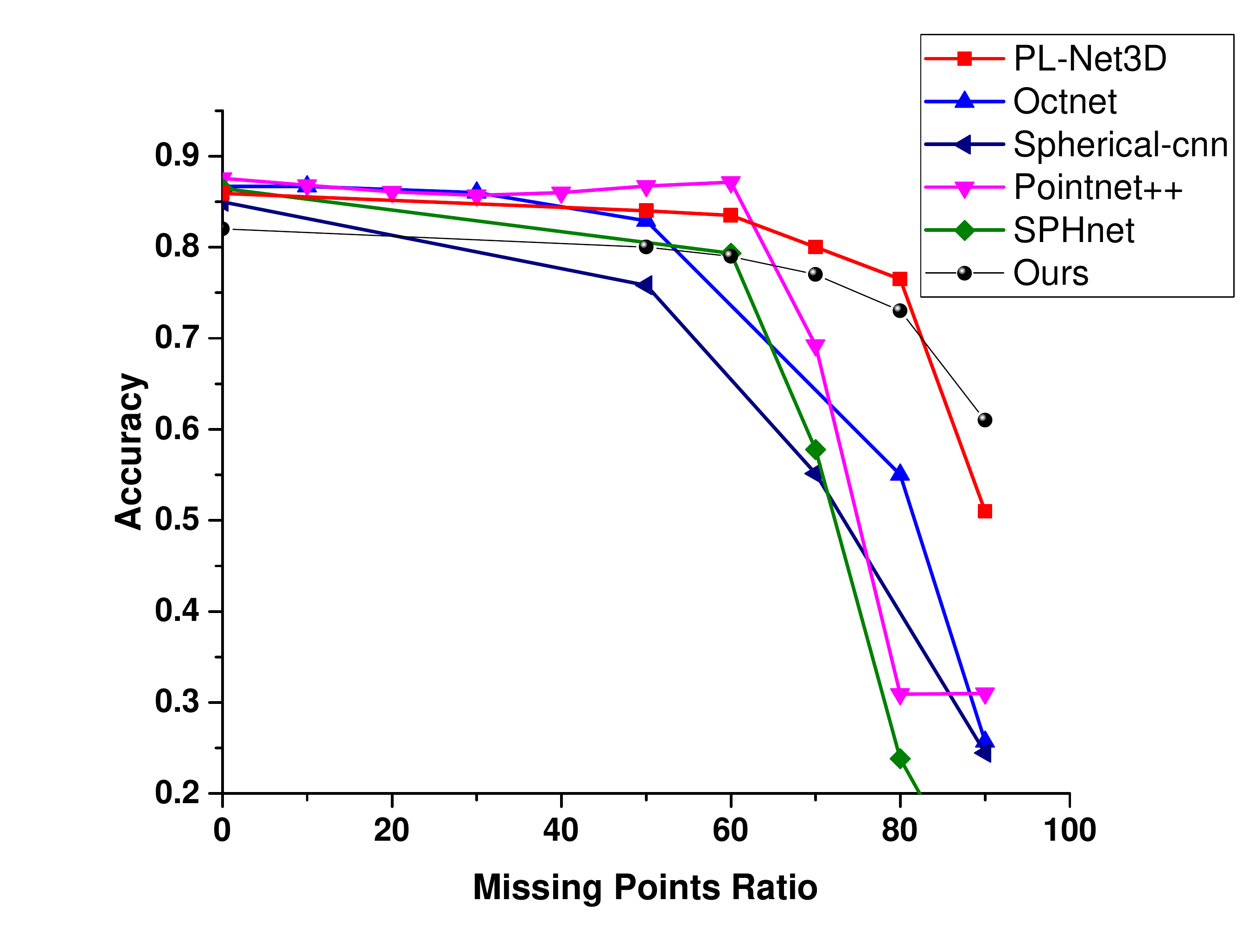}
\caption{Classification accuracy versus missing points.}
\label{fig_missp}
\end{figure}

We also evaluated the effect of missing points on the classification performance of different methods. Figure~\ref{fig_missp} shows that our model classification accuracy drops by only 2\% when half the points are eliminated, and by 20\% when 90\% of points are removed. spherical-cnn classification accuracy drops by 10\% when half the points are eliminated and it degrades after that. SPHnet classification accuracy drops by 8\% when 60\% of points are eliminated and it degrades after that. Ocntnet classification accuracy degrades after 50\%.

\section{Ablation study} \label{sec:ABLATION STUDY}

We conducted a number of experiments to explore all possible solutions of our method and their performance under different data augmentation and the results are shown in table~\ref{Tab:comp}. The first row shows the results of our proposed model (RSCNN) with 3D Models having 2000 points (same results shown in the previous section), while the second and the rest of the raw's show the results when training without noise addition. We implemented the inverse transform operation after applying the convolution in our model. As a result, our model performed worse and became less robust to data augmentation.  The outputs are presented in the third row of table~\ref{Tab:comp}. The effect of performing Inverse Fourier Transform on the classification accuracy is discussed in the Appendix. 
In the next step, we evaluated our model with no fully connected layer to reduce the number of trainable parameters. However, the results shown in the third row, suggest that such an action is detrimental for the overall performance. 

As most existing techniques used only 2000 points in their experiments, we also used the same number of points in our comparative experiments. However, for objects with 10,000 points, as shown in the forth row, our network achieves higher performance and robustness. The classification accuracy with 50\% outliers increased to 79\%, while we noticed 1\% enhancement for both clean and noisy objects. 
When testing Spherical-cnn~\cite{esteves2018learning} utilizing the same grid we used, the performance of their method enhanced tremendously. Although spherical-cnn was able to achieve 70\% classification accuracy with 50\% outliers, it is still around 10 percent lower than the performance of the proposed method. 


\begin{table}[]
\caption{Classification accuracy versus different network architectures and different data augmentations}
\label{Tab:comp}
\fontsize{10}{12}\selectfont 

\begin{tabular}{|l|c|c|c|c|c|}
\hline
\multicolumn{1}{|c|}{\multirow{2}{*}{method}} & \multirow{2}{*}{\begin{tabular}[c]{@{}c@{}}\\  sampled\\  points\end{tabular}} & \multicolumn{4}{c|}{classification accuracy}                                                                                                                                         \\ \cline{3-6} 
\multicolumn{1}{|c|}{}                        &                                                                                       & clean & \begin{tabular}[c]{@{}c@{}}80\% \\  dropout\end{tabular} & \begin{tabular}[c]{@{}c@{}}0.1\\ noise\end{tabular} & \begin{tabular}[c]{@{}c@{}}50\%\\ outliers\end{tabular} \\ \hline
ours                                          & 2000                                                                                  & \textbf{0.82}  & 0.72                                                           & 0.62                                                & 0.74                                                    \\ \hline

ours*                                        & 2000                                                                                  & 0.81  & 0.72                                                           & 0.63                                                & 0.75                                                    \\ \hline

ours + IFT                                    & 2000                                                                                  & 0.78  & 0.71                                                           & 0.45                                                & 0.58                                                    \\ \hline
ours no FC                                    & 2000                                                                                  & 0.75  & 0.65                                                           & 0.54                                                & 0.46                                                    \\ \hline
\multicolumn{6}{|l|}{}                                                                                                                                                                                                                                                                                                         \\[-2ex] \hline
ours                                          & 10000                                                                                 & \textbf{0.82}  & \textbf{0.80}                                                            & \textbf{0.64}                                                & \textbf{0.79}                                                    \\ \hline
sph-cnn \cite{esteves2018learning}*                                & 10000                                                                                 & 0.80   & 0.53                                                           & 0.43                                                & 0.69                                                    \\ \hline
\end{tabular}
\end{table}

\subsection{Grid resolution}

\begin{figure}
\centering
\includegraphics[scale=.3]{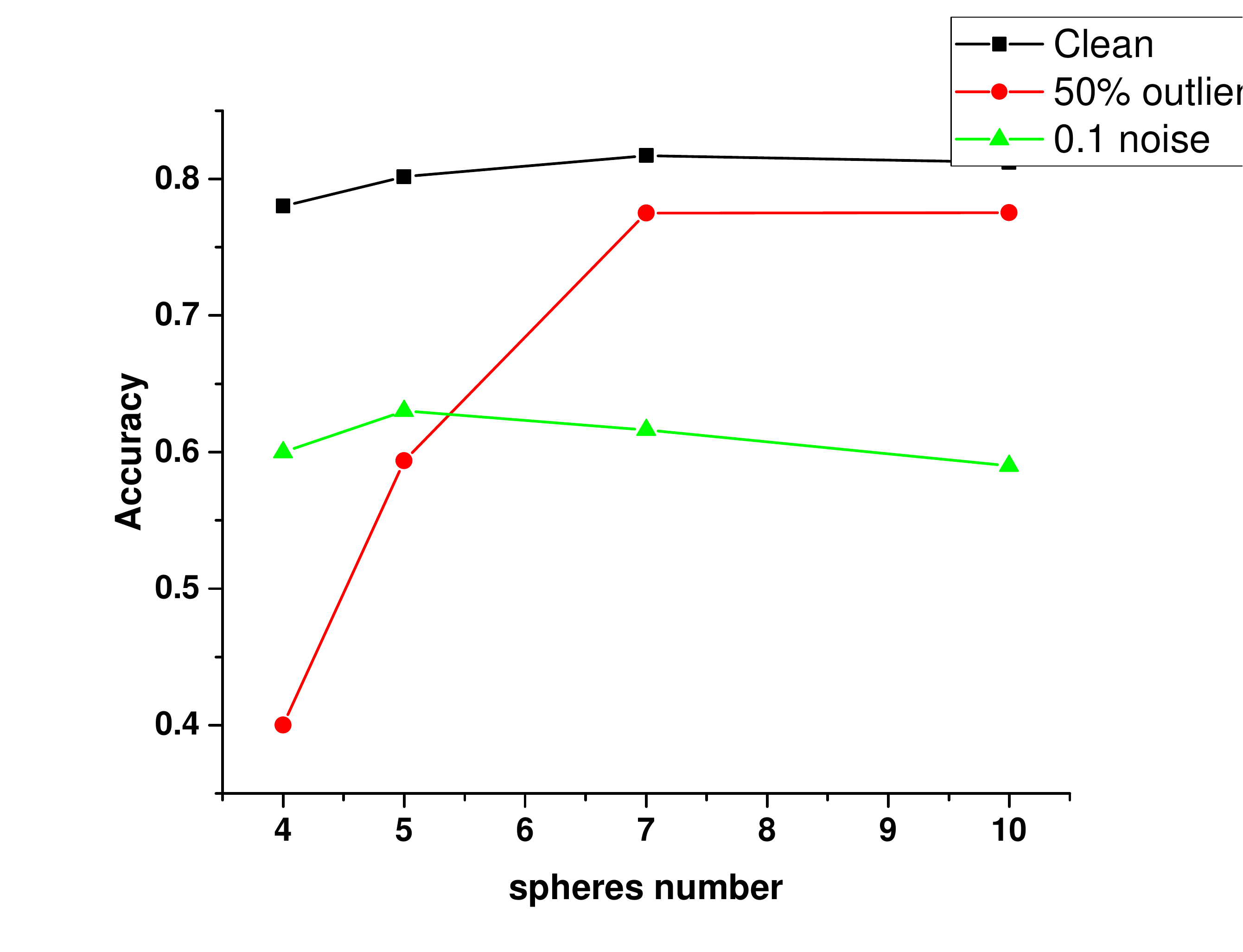}
\caption{Classification accuracy versus number of concentric spheres.}
\label{fig_c}
\end{figure}

We tested our method with 4, 5, 7, and 10 concentric spheres where each sphere has a 64 by 64 grids. The results are shown in Figure~\ref{fig_c}. For both clean and outlier-augmented objects, our network performance gradually increases up to using 7 concentric spheres and plateaus afterwards. As such, we used only 7 concentric spheres in our experiments. 

\subsection{Number of layers}

\begin{figure}[]
\centering
\includegraphics[scale=.33]{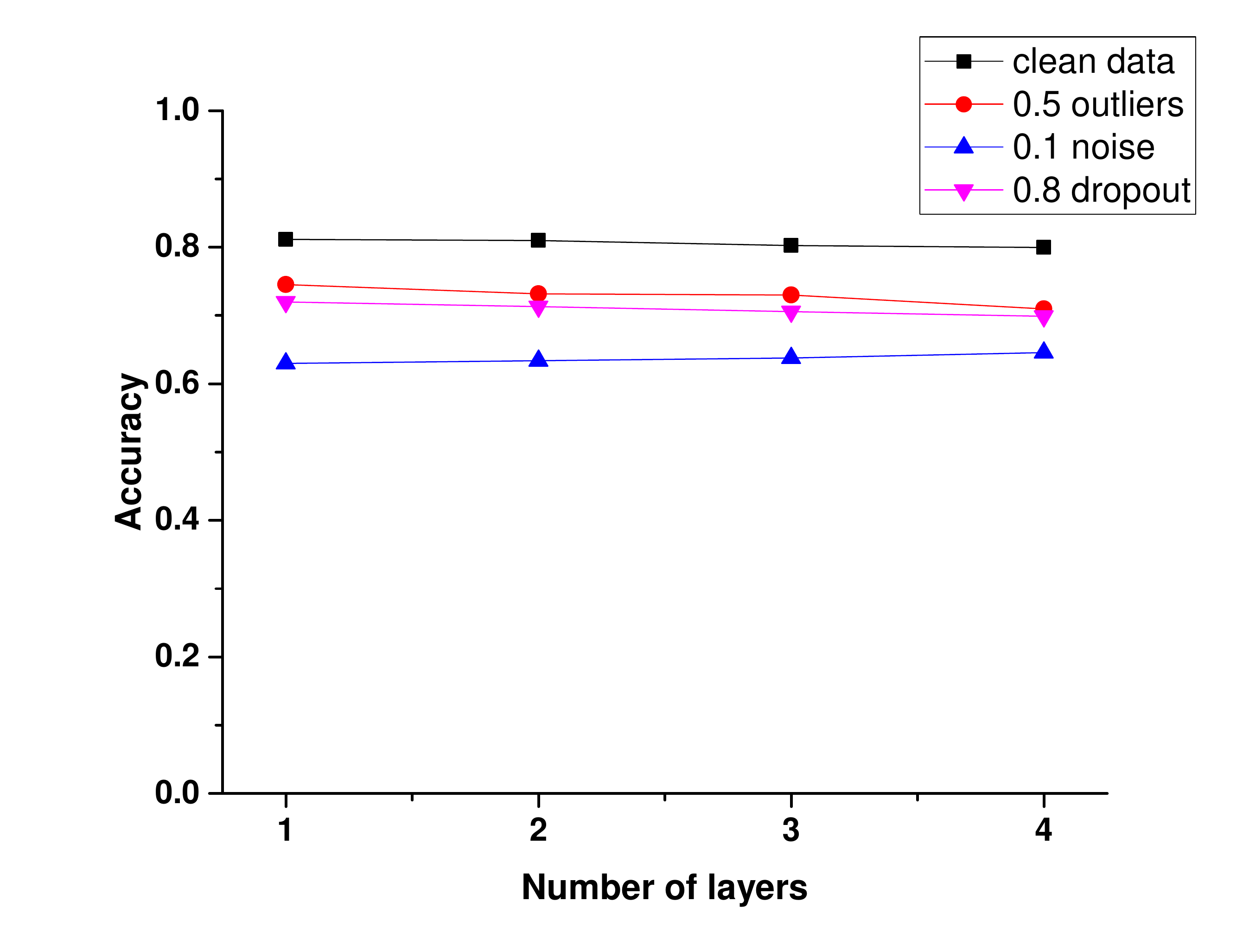}
\caption{Classification accuracy versus number of layers for our network.}
\label{fig_l}
\end{figure}

We tested with different architectures having one to four layers as shown in Figure~\ref{fig_l}. We find that increasing the number of layers from 1 to 4 does not improves the accuracy on clean and augmented objects, thus we restricted our model to a single convolutional layer.

\section{Limitation and future work}

 In theory, we expect our model to achieve better performance compared to Spherical-cnn~\cite{esteves2018learning} as we learn features from the unit ball. However, Spherical-cnn~\cite{esteves2018learning} was able to achieve higher classification accuracy for clean objects. Nevertheless, The proposed framework performance is competitive to other robust methods that exist in the literature \cite{mukhaimar2019pl,gould2019deep}. 
 

\section{Conclusion}  \label{sec:Conclusion}

Spherical harmonics have been used for object classification over the years with several frameworks existing in the literature. Their rotation equavarience, and the reduced number of trainable parameters are key properties they exhibit. However, their robustness to data augmentation have not been studied before.
Point clouds of 3D models are often imperfect and contain outliers. Therefore, it is important to use a robust classification framework that can overcome such inaccuracies. 
In this paper we present the robustness of spherical harmonics approaches available in the literature for classification of point cloud objects. We also propose a robust model that have shown to compete with the state-of-art models. Our model uses the voxel grid of concentric spheres to learn features over the unit ball. In addition, we keep the convolution operations in the Fourier domain without applying the inverse transform used in previous approaches. As a result, our model is able to learned features that are less sensitive to data augmentation. We tested our proposed model against several types of data augmentation such as noise and outliers. Our results show that the proposed model outperforms the state of art networks in terms of robustness to effects of data augmentation.  

\appendices
\section{Effect of inverse Fourier transform }

\begin{figure*}[]
\centering
\includegraphics[scale=.7]{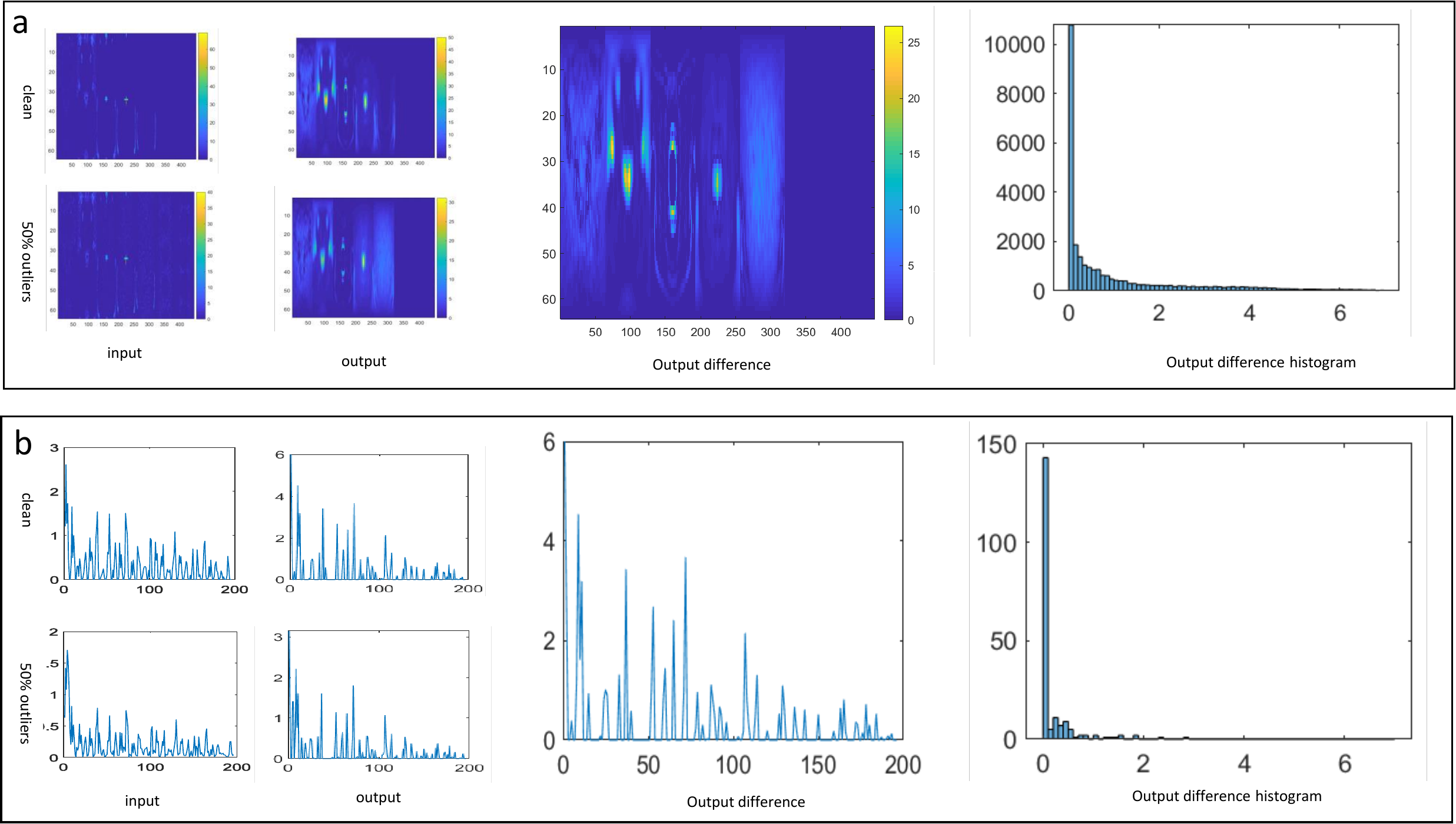}
\caption{A comparison between the spherical convolution operation with (a) IFT applied after the convolution, and (b) no IFT applied. In both scenarios, we have a clean shape and the same shape augmented with outliers.  The first column is the input signal, where in (a) we have a voxel grid signal that consists of 64*64*7 voxels (spherical grid). The color bar represents the number of points inside each voxel. while the first column in (b) represents the spherical harmonics coefficients of the same input signal shown in (a). The second column shows the output of the spherical convolution operation, where in (a) we apply IFT, hence, the signal is in the spherical grid domain, while in (b) the signal is kept in the Fourier domain. The third column shows the difference in signal between the clean and outlier case, and the last column shows the histograms of the images in the third column. Histograms have the same bin width.}
\label{fig_case}
\end{figure*}

Our experiments show that applying inverse Fourier transform (IFT) reduces robustness to outliers by almost 15 percent. To demonstrate this, we compared the convolution operation of both cases in Figure~\ref{fig_case}, where~(a) represents the convolution operation that uses IFT, and~(b) represents our convolution operation. In Figure~\ref{fig_case}-(a), we plot the input voxel grid as a 2D image having a dimension of 64 by 64*7, where 7 is the number of concentric spheres. The pixel values represent the values of the voxel grids. The color bar shows how many points inside each voxel.  While in Figure~\ref{fig_case}-(b), we plot the Fourier coefficients instead of the voxel grid. For each case, a signal of a clean shape and a signal of the same shape having outliers are used as inputs. Next we apply the convolution and plot their outputs in the second column. Since in (a) the convolution uses IFT, the output signal is again returned to the Euclidian domain. However for our convolution in (b), our output signal is kept in the Fourier domain.
Finally, we plot the difference between both outputs in the third column with its associated histogram in the last column. Comparing the histograms of (a) and (b) show that our convolution operation produces less error.

\section{t-SNE}
\begin{figure*}[]
\centering
\includegraphics[scale=.7]{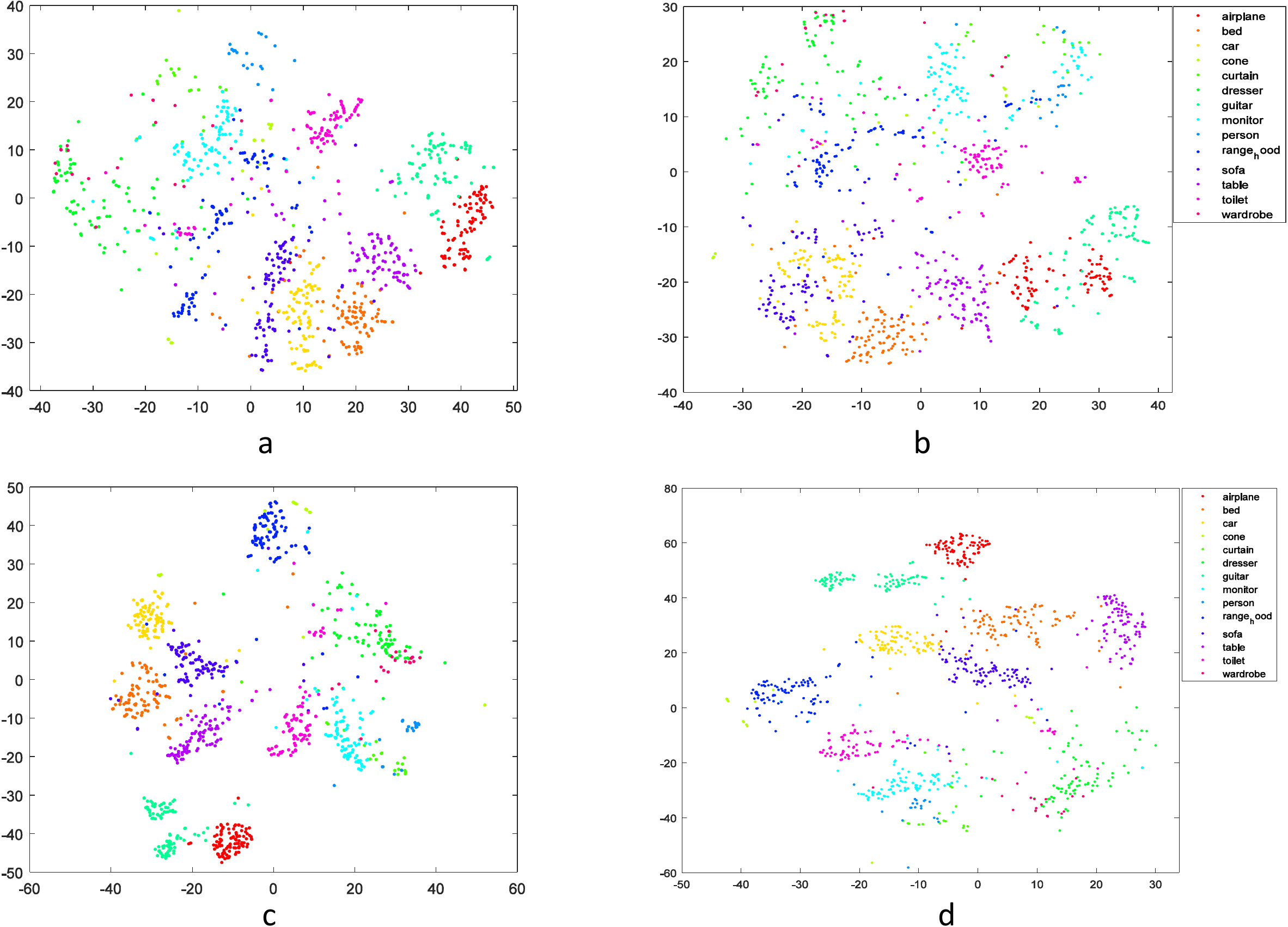}
\caption{Visualization of 2D embedding of shape features using t-SNE technique for: (a) a simple classifier using clean data, (b) same classifier  using 50\% outliers augmented data, (c) our model on clean data, and (d) our model on 50\% outliers augmented data. }
\label{fig_tsne}
\end{figure*}

We used the t-Distributed Stochastic Neighbor Embedding (t-
SNE) \cite{maaten2008visualizing} to visualize 2D embedding of learnt shape features in figure~\ref{fig_tsne}. We only show a few representative labels results. The t-sne technique helps us to visualizes high-dimensional feature vectors (1024 dimensions) aggregated from the convolution operation into the 2D space shown in the figure. Figure~\ref{fig_tsne}-(a,b) show the t-sne embedding for a simple classifier that uses equation~(\ref{equ:desc1}) for classification.  We show scenarios of clean data of ModelNet40 test split, and the 50\% outlier augmented data respectively. Similarly in figure~\ref{fig_tsne}-(c,d) for our model.
The results show that using our model helped in better grouping similar labels for clean data and augmented data as well. We still can see the labels clustered clearly in the presence of outliers, which means that our learned feature is almost not affected by the presence of outliers.

\ifCLASSOPTIONcaptionsoff
  \newpage
\fi



%
\bibliographystyle{IEEEtran} 

\bibliography{ref}

%




\end{document}